# Recognition of Handwritten Textual Annotations using Tesseract Open Source OCR Engine for *information Just In Time (iJIT)*

Sandip Rakshit [1], Subhadip Basu [2], Hisashi Ikeda [3]

[1] Techno India College of Technology, Kolkata, India
[2] Computer Science and Engineering Department, Jadavpur University, India
[3] Intelligent Media Systems Department, Central Research Laboratoty, Hitachi Limited, Japan

[1] Corresponding author. E-mail: subhadip@ieee.org

## Abstract

Objective of the current work is to develop an Optical Character Recognition (OCR) engine for information Just In Time (iJIT) system that can be used for recognition of handwritten textual annotations of lower case Roman script. Tesseract open source OCR engine under Apache License 2.0 is used to develop user-specific handwriting recognition models, viz., the language sets, for the said system, where each user is identified by a unique identification tag associated with the digital pen. To generate the language set for any user, Tesseract is trained with labeled handwritten data samples of isolated and free-flow texts of Roman script, collected exclusively from that user. The designed system is tested on five different language sets with free- flow handwritten annotations as test samples. The system could successfully segment and subsequently recognize 87.92%, 81.53%, 92.88%, 86.75% and 90.80% handwritten characters in the test samples of five different users.

## 1. Introduction

In online character recognition, the trajectories of pen tip movements are recorded and analyzed to identify the linguistic information expressed. With the latest technological advancements in pen input devices, new interfaces are designed to capture the precise pen-trajectory information and subsequent analysis of online handwritten data, with user comforts in writing. It is now possible to write on an ordinary paper and immediate wireless transmission of handwritten annotations to a remote server [1]. With these technological advances, handwritten annotations in digital notebooks may be digitized in no time. Traditionally, documents containing handwritten information are difficult to archive in digital form. Even with the help of latest optical scanners, content based indexing techniques and research tools; it is difficult to find digitized versions of document pages based on user queries. Some work has recently been done on content based retrieval of handwritten documents [2-4]. In [2], Bertrand et.al. have developed a technique for structural document recognition and recognition of handwritten names. In another work, Matthew et.al. [3] developed a stroke feature based technique for retrieval of handwritten Chinese annotations based on typed/handwritten query. Srihari et.al. [4] had used stroke/shape features for indexing and retrieval of handwritten documents based on writer characteristics, textual content and writer profile. In one of our earlier works [5], a recognition based indexing technique was discussed for real-time retrieval of handwritten annotations based on typed/handwritten query. David Doermann, in his survey [6], had highlighted key issues involved in indexing and retrieval of document images.

In any recognition based indexing technique, the overall performance predominantly depends on accuracy of the underlying recognition engine. Development of a handwritten OCR engine with high recognition accuracy is a still an open problem for the research community. Lot of research efforts have already been reported [7-9] on different key aspects of handwritten character recognition systems. In this work, we have used Tesseract 2.01 [10], an open source OCR Engine under Apache License 2.0, for segmentation and subsequent recognition of handwritten textual annotations of lower case of *Roman* script.

Objective of the current work is to develop an Optical Character Recognition (OCR) engine for the *iJIT* system that can be used for recognition of handwritten textual annotations of lower case *Roman* script. Tesseract is used to develop user-specific handwriting recognition models, *viz*., the language sets, for the said system. Each user of the *iJIT* system may be identified by a unique identification tag associated with the *Anoto* digital pen [1]. Tesseract OCR engine is customized to perform user specific training on labeled handwriting samples of both isolated and free-flow texts, written using lower case *Roman* script. The performance is evaluated on both the categories of document pages for observation of segmentation and character recognition accuracies.

The following sections describe an overview of the existing *iJIT* system, an overview of the Tesseract OCR engine and the present experiment on designing an OCR engine for segmentation and recognition of handwritten textual annotations.

## 2. The *iJIT* system

Just in time availability of meaningful information is the key to any real-time information retrieval system. The *information Just In Time (iJIT)* system [5], developed at the Hitachi Central Research Laboratory, keeps track of all the digital documents stored in the *iJIT* server. Using the proposed system, handwritten annotations on the printed digital document pages using *Anoto* digital pens [1] may be viewed/shared/searched based on typed/handwritten query. Fig. 1. shows a schematic overview of the recognition based query retrieval scheme designed for the *iJIT* system.

The *iJIT* system uses ordinary papers, attached with digitally legible patent-protected dot-patterns from *Anoto* [1], for printout of each digital document through the server. The *Anoto* patterns consist of numerous nearly invisible, intelligent black dots that can be read by a digital pen. The pattern on each paper is unique so that each page can be kept separate from one another.

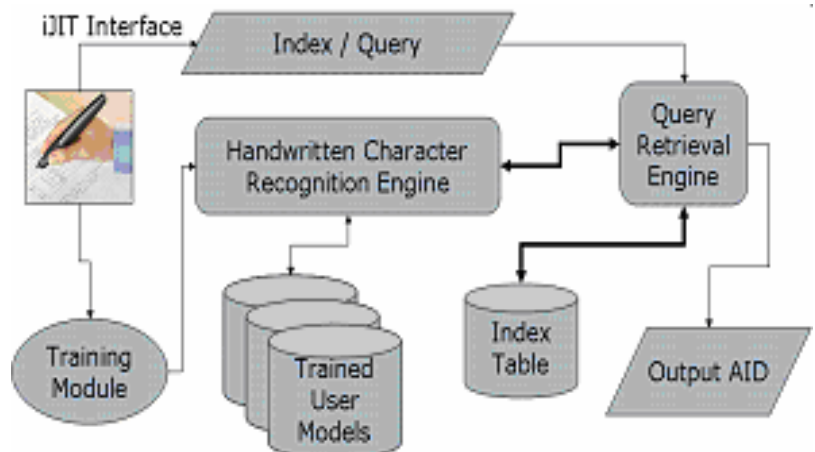

Fig. 1. A schematic architecture of the recognition based query retrieval system.

An *Anoto* digital pen [1], looks like its normal ballpoint counterpart, contains an integrated digital camera, an advanced image microprocessor and a wireless communication device. The pen can take around 50 digital snapshots per second, can store up to 50 full A4/letter size pages of handwritten data and then securely send the information to the *iJIT* server through wireless communication or *Universal Serial Bus (USB)*. Every snapshot contains enough data to determine the exact position of the pen in the paper, the time of pen-stroke and the unique identification number of the *Anoto* paper. Each pen is also having unique identification numbers so that the *iJIT* system can distinguish between every individual's handwriting.

## 3. Overview of the Tesseract OCR engine

Tesseract is an open source (under Apache License 2.0) offline optical character recognition engine, originally developed at Hewlett Packard from 1984 to 1994. Tesseract is now partially funded by Google [10] and released under the Apache license, version 2.0. The latest version, Tesseract 2.03 is released in April, 2008. In the current work, we have used Tesseract version 2.01, released in August 2007.

Like any standard OCR engine, Tesseract is developed on top of the key functional modules like, line and word finder, word recognizer, static character classifier, linguistic analyzer and an adaptive classifier. However, it does not support document layout analysis, output formatting and graphical user interface. Currently, Tesseract can recognize printed text written in English, Spanish, French, Italian, Dutch, German and various other languages.

To train Tesseract in English language 8 data files are required in tessdata sub directory. The 8 files used for English are to be generated as follows:

tessdata/eng.freq-dawg
tessdata/eng.word-dawg
tessdata/eng.user-words
tessdata/eng.inttemp
tessdata/eng.normproto
tessdata/eng.pffmtable
tessdata/eng.unicharset
tessdata/eng.DangAmbigs

## 4. The present work

In the current work, Tesseract 2.01 is used for developing user-specific handwriting recognition models, *viz.*, the language sets, for the *iJIT* system. To generate the language sets for each user, Tesseract is trained with labeled handwritten data samples of isolated and free-flow texts of lower case *Roman* script. Key functional modules of the developed system are discussed in the following sub-sections.

### 4.1. Collection of the dataset

For preparation of the dataset for the current experiment, digitized handwritten samples of lower case *Roman* script were collected from five different users. Six handwritten document pages, consisting of isolated characters and free-flow words were collected from each of the users of the designed system. These pages are categorized into two datasets. Dataset-1 consists of four pages of isolated handwritten lower case *Roman* characters and Dataset-2 constitutes two pages of free-flow handwritten words, written from technical articles. For each user, three pages from the dataset-1 and one page from the dataset-2 were considered for training the Tesseract OCR engine. The remaining two pages, one from each dataset, constitute the test set for the current experiment. The overall distribution of the character samples in the training and the test sets for the five users is shown in Table 1.

Table 1. Composition of the training and test set character samples for different users

|  |  | Dataset-1 | Dataset-2 | Overall |
|---|---|---|---|---|
| User 1 | Train set | 1185 | 659 | 1844 |
|  | Test set | 442 | 691 | 1133 |
| User 2 | Train set | 1006 | 529 | 1535 |
|  | Test set | 468 | 718 | 1186 |
| User 3 | Train set | 992 | 884 | 1876 |
|  | Test set | 546 | 1004 | 1550 |
| User 4 | Train set | 619 | 578 | 1197 |
|  | Test set | 260 | 751 | 1011 |
| User 5 | Train set | 467 | 255 | 722 |
|  | Test set | 234 | 277 | 511 |

## 4.2. Labeling training data

For labeling the training samples of each user using Tesseract we have taken help of a tool named bbTesseract [13]. To generate the training files for a specific user, we need to prepare the box files for each training images using the following command:

*tesseract fontfile.tif fontfile batch.nochop makebox*

The box file is a text file that includes the characters in the training image, in order, one per line, with the coordinates of the bounding box around the image. Incorrect labels in the training set may be manually corrected using the bbTesseract Tool.

Then we have to rename the boxfile fontfile.txt to fontfile.box. Fig. 2 shows a screenshot of the bbTesseract tool.

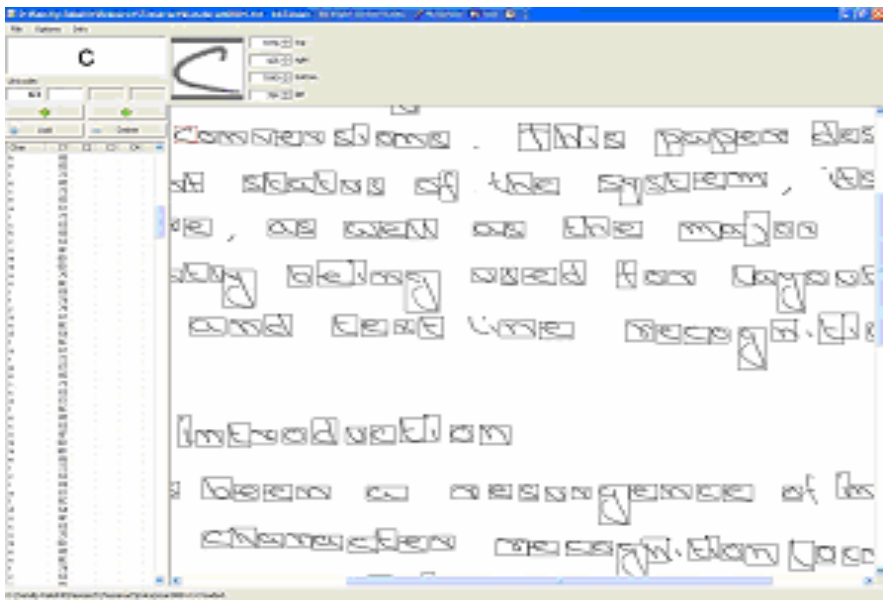

Fig.2. A sample screenshot of a segmented training page using the bbTesseract tool

### 4.3. Training the data using Tesseract OCR engine

For training a new language set for any user, we have to put in the effort to get one good box file for a handwritten document page, run the rest of the training process, discussed below, to create a new language set. Then use Tesseract again using the newly created language set to label the rest of the box files corresponding to the remaining training images using the process discussed in section 4.2.

For each of our training image, boxfile pairs, run Tesseract in training mode using the following command:

*tesseract fontfile.tif junk nobatch box.train*

The output of this step is fontfile.tr which contains the features of each character of the training page. The character shape features can be clustered using the mftraining and cntraining programs:

*mftraining fontfile_1.tr fontfile_2.tr ...*

This will output three data files: inttemp , pffmtable and Microfeat, and the following command:

*cntraining fontfile_1.tr fontfile_2.tr ...*

This will output the normproto data file. Now, to generate the unicharset data file, unicharset_extractor program is used as follows:

*unicharset_extractor fontfile_1.box fontfile_2.box ...*

Tesseract uses 3 dictionary files for each language. Two of the files are coded as a Directed Acyclic Word Graph (DAWG), and the other is a plain UTF-8 text file. The wordlist is formatted as a UTF-8 text file with one word per line. The corresponding commands are:

*wordlist2dawg frequent_words_list freq-dawg*

*wordlist2dawg words_list word-dawg*

The third dictionary file name is user-words and is usually empty. The final data file of Tesseract is DangAmbigs file. This file cannot be used to translate characters from one set to another. The DangAmbigs file may be empty also.

Now we have to collect all the 8 files and rename them with a lang. prefix, where lang is the 3-letter code for our language and put them in our tessdata directory. Tesseract can then recognize text in our language set using the command:

*tesseract image.tif output -l lang*

## 5. Experimental results

For conducting the current experiment, five user-specific language sets are generated using Tesseract open source OCR engine. The training and test patterns of each individual user are spread over two types of datasets, as described in Sec. 4.1. The experiment is focused on testing the segmentation and core recognition accuracy of Tesseract OCR engine on free flow handwritten annotations written using digital pens by different users. The linguistic analysis module of Tesseract, involving the language files freq-dawg, word-dawg, user-words and DangAmbigs are not utilized in the current experiment. To evaluate the performance of the present technique the following expression is developed.

Recognition accuracy = $(CB_{tB} / (CB_{mB} + CB_{sB}))*100$

where $CB_{tB}$ = the number of character segments producing true classification result and $CB_{mB}$ = the number of misclassified character segments and $CB_{sB}$ signifies the number of character Tesseract fails to segment, i.e., producing under segmentation. The rejected character/word samples are excluded from computation of recognition accuracy of the designed system.

Table 2(a-e) shows an analysis of successful classification (SC), misclassification (Misc), segmentation failure (SF) and rejection (Rej) results on the test samples of the three users. Fig. 3 shows a character wise distribution of success and failure accuracies on the overall test dataset. As observed from the experimentation a significant proportion rejection cases evolve out of the word segmentation failures. This is so because Tesseract is originally designed to recognize printed document pages with uniformity in baseline and character/word spacings. Another source of error is due to the internal segmentation of some of the characters. More specifically, the character 'i' often gets internally segmented into two parts, leading to high individual error rates.

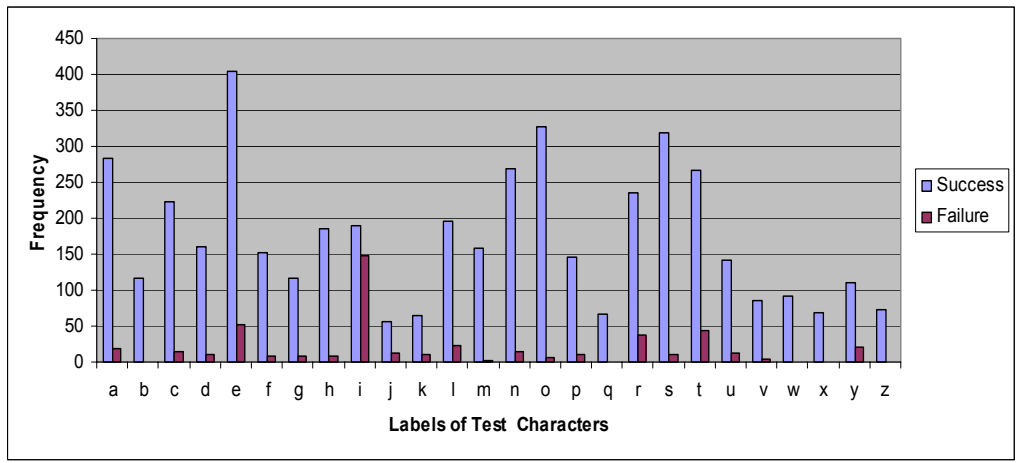

Fig. 3. Distribution of success and failure cases over the free flow test page.

Table 2. Analysis of recognition performance of the developed system

(a) Recognition performance of User-1 test dataset

|  | Dataset-1 | Dataset-2 | Overall |
|---|---|---|---|
| **SC** | 95.42 | 83.2 | 87.92 |
| **Misc** | 4.1 | 16.19 | 11.52 |
| **SF** | 0.48 | 0.61 | 0.56 |
| **Rej** | 6.10 | 4.34 | 5.03 |

(b) Recognition performance of User-2 test dataset

|      | Dataset-1 | Dataset-2 | Overall |
|------|-----------|-----------|---------|
| SC   | 91.62     | 76.45     | 81.53   |
| Misc | 8.38      | 18.31     | 15.00   |
| SF   | 0.00      | 5.24      | 3.47    |
| Rej  | 26.07     | 4.18      | 12.82   |

(c) Recognition performance of User-3 test dataset

|      | Dataset-1 | Dataset-2 | Overall |
|------|-----------|-----------|---------|
| SC   | 96.78     | 90.94     | 92.88   |
| Misc | 3.22      | 6.18      | 5.19    |
| SF   | 0.00      | 2.88      | 1.93    |
| Rej  | 8.97      | 0.00      | 3.16    |

(d) Recognition performance of User-4 test dataset

|      | Dataset-1 | Dataset-2 | Overall |
|------|-----------|-----------|---------|
| SC   | 90.38     | 85.49     | 86.75   |
| Misc | 8.85      | 7.32      | 7.72    |
| SF   | 0.77      | 7.19      | 6.03    |
| Rej  | 0         | 0         | 0       |

e) Recognition performance of User-5 test dataset

|      | Dataset-1 | Dataset-2 | Overall |
|------|-----------|-----------|---------|
| SC   | 91.88     | 89.89     | 90.80   |
| Misc | 8.12      | 10.11     | 9.20    |
| SF   | 0         | 0         | 0       |
| Rej  | 0         | 0         | 0       |

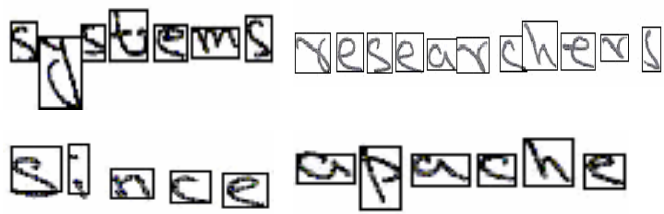

Fig. 4. Some of the successfully segmented and recognized word images.

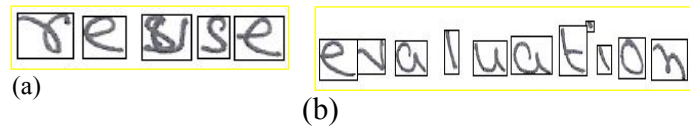

(a)      (b)

Fig. 5. Some of the misclassified word images

(a) Recognition error in the 3[rd] character
(b) Internal segmentation in the 8[th] character

As shown in Table 2(a-e), the overall character-level recognition accuracy of the developed system is around 87.98%. The overall character misclassification rate is observed as around 9.73%. Segmentation failures in the document pages account for around 2.29% error cases. The reason behind high segmentation failure is due to the over-segmentation of some of the constituent characters like 'i', 'j' and also due to under-segmentation of cursive words in the document pages. The designed system rejects around 9.24% characters in the test dataset. This is mainly due to the presence of multi-skewed handwritten text lines in the test documents. Completely cursive words were also rejected completely in many cases during the experimentation. Some of the sample word images successfully segmented and recognized by Tesseract are shown in Fig. 4. Fig. 5(a-b) shows some of the word images with erroneous segmentation and recognition results.

A major drawback of the current system is its failure to avoid over-segmentation in some of the characters. Also the system fails to segment cursive words in many cases leading to under-segmentation and rejection. The recognition performance of the designed system may further be improved by incorporating more training samples for each user and inclusion of word-level dictionary matching techniques.

Despite these limitations, the designed recognition engine is successfully integrated with the *iJIT* system for online interpretation of handwritten textual annotations. The word-level recognition time of the OCR engine, as observed on reasonably powered computer hardware, is also found to be satisfactory. In a nutshell, the current work effectively customizes an open source OCR engine for segmentation and recognition of handwritten textual annotations of multiple users within the designed *iJIT* system.